\pdfoutput=1
\documentclass[11pt]{article}

\usepackage{acl}
\usepackage{times}
\usepackage{latexsym}

\usepackage[T1]{fontenc}
\usepackage[utf8]{inputenc}

\usepackage{algorithm}
\usepackage{microtype}
\usepackage{algcompatible,amsmath}
\usepackage{graphicx}
\usepackage{url}
\usepackage{multirow}
\usepackage{comment}
\usepackage{amssymb}
\usepackage{pifont}
\usepackage{balance}
\usepackage{mathtools}
\usepackage{xcolor}
\usepackage{colortbl}
\usepackage{bbm}
\usepackage{enumitem}
\usepackage{marvosym}
\usepackage{textcomp}
\usepackage{array}
\usepackage{booktabs}
\usepackage{tabularx}
\usepackage{CJKutf8}

\usepackage{newfloat}
\usepackage{listings}
\floatstyle{ruled}
\newfloat{listing}{tb}{lst}{}
\floatname{listing}{Listing}

\newcommand{\squishlist}{
\begin{list}{$\bullet$}
{ \usecounter{Lcount}
\setlength{\itemsep}{0pt}
\setlength{\parsep}{0pt}
\setlength{\topsep}{0pt}
\setlength{\partopsep}{0pt}
\setlength{\leftmargin}{2em}
\setlength{\labelwidth}{1.5em}
\setlength{\labelsep}{0.5em} } }
\newcommand{\squishend}{
\end{list} }

\title{Scene Graph Modification as Incremental Structure Expanding}

\author{Xuming Hu$^{1*}$, Zhijiang Guo$^{2*}$, Yu Fu$^1$, Lijie Wen$^{1\dagger}$, Philip S. Yu$^{1,3}$\\
  $^1$Tsinghua University\\  $^2$University of Cambridge\\ $^3$University of Illinois at Chicago\\
  $^1$\texttt{\{hxm19,fy20\}@mails.tsinghua.edu.cn}\\
  $^2$\texttt{zg283@cam.ac.uk}
  $^1$\texttt{wenlj@tsinghua.edu.cn}
  $^3$\texttt{psyu@uic.edu}\\
  }

\begin{document}
\maketitle
\begin{abstract}
A scene graph is a semantic representation that expresses the objects, attributes, and relationships between objects in a scene. Scene graphs play an important role in many cross modality tasks, as they are able to capture the interactions between images and texts. In this paper, we focus on scene graph modification (SGM), where the system is required to learn how to update an existing scene graph based on a natural language query. Unlike previous approaches that rebuilt the entire scene graph, we frame SGM as a graph expansion task by introducing the incremental structure expanding (ISE). ISE constructs the target graph by incrementally expanding the source graph without changing the unmodified structure. Based on ISE, we further propose a model that iterates between nodes prediction and edges prediction, inferring more accurate and harmonious expansion decisions progressively. In addition, we construct a challenging dataset that contains more complicated queries and larger scene graphs than existing datasets. Experiments on four benchmarks demonstrate the effectiveness of our approach, which surpasses the previous state-of-the-art model by large margins. Source code and data are available\footnote{\url{https://github.com/THU-BPM/SGM}\\\phantom{00} $^{*}$ Equally Contributed.\\\phantom{00} $^\dagger$ Corresponding Author.}.
\end{abstract}
\section{Introduction}
\label{sec:intro}
\begin{figure}
    \centering
    \includegraphics[scale=0.4]{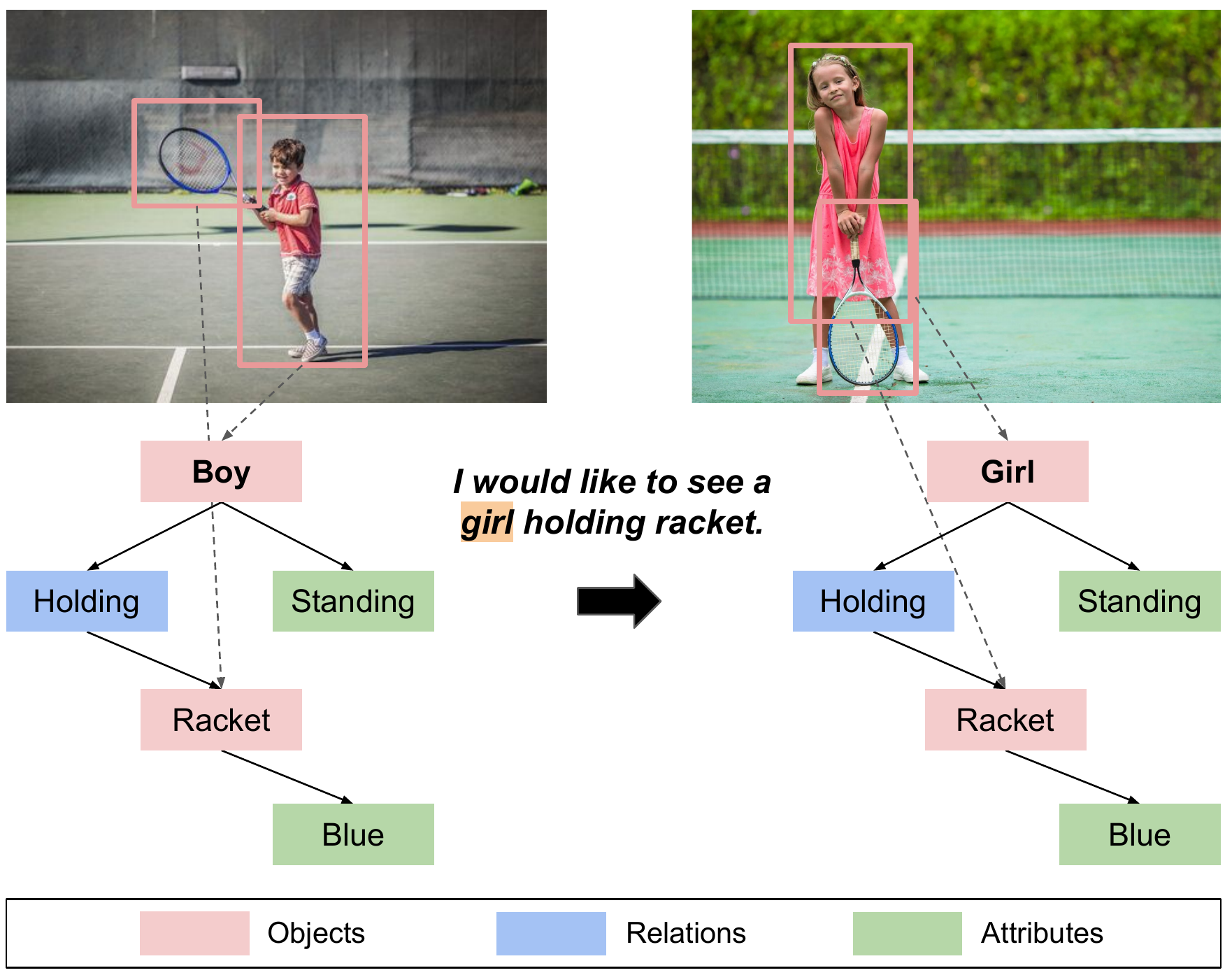}
    % \vspace{-1em}
    \caption{Example images and their corresponding scene graphs. Given the query, the original scene graph (left) is modified to be the target scene graph (right). }
    \label{fig:example}
    % \vspace{-1em}
\end{figure}
A scene graph is a structural representation that captures the semantics of visual scenes by encoding object instances, attributes of objects, and relationships between objects.
~\citep{Johnson2015ImageRU}. As shown in Figure~\ref{fig:example}, the scene graph encodes objects (e.g.\ ``\textit{Boy}'', ``\textit{Racket}''), attributes (e.g.\ ``\textit{Girl is standing}''), and relations (``\textit{Boy holding racket}''). Scene graphs are able to capture the interactions between text and images by associating objects in the graph with regions of an image and modeling the relations between objects. Therefore, it has been used in the cross modality task such as image retrieval, image captioning, and visual question answering~\citep{Schuster2015GeneratingSP, shi2019explainable, YangTZC19, Wang2020CrossmodalSG}. 
\begin{figure*}
    \centering
    \includegraphics[scale=0.63]{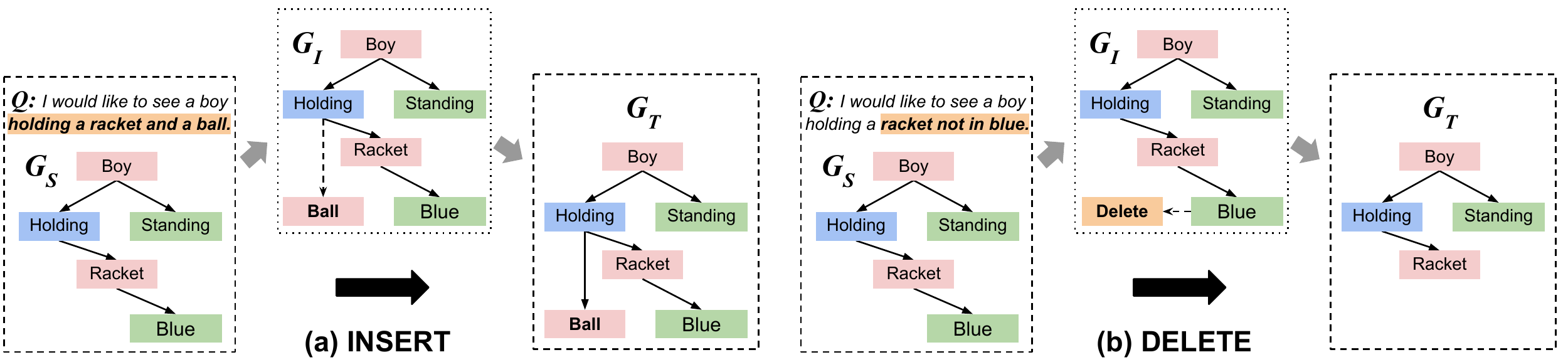}
    % \vspace{-1em}
    \caption{Examples of basic operations INSERT and DELETE for scene graph modification. $Q$ denotes the textual query, $G_{S}$ denotes the source scene graph, $G_{T}$ denotes the target scene graph and $G_{I}$ is the extended graph. }
    \label{fig:ise}
    % \vspace{-1em}
\end{figure*}

Recently, modifying the scene graph based on the input becomes an emerging research direction as cross-modal systems may need to resort to an interactive process through multiple iterations~\citep{Ramnath2019SceneGB,He2020SceneGM}. Take text-based image retrieval as an example, users start with a query describing the main objects or topics they are looking for, then modify the query to add more constraints or provide additional information based on previous search results. Instead of directly manipulating images, scene graphs can be used to convert the image-editing problem into a graph-editing problem, conditioned on the textual query. As shown in Figure~\ref{fig:example}, given a retrieved image from the last turn, if the user wants to see a girl rather than a boy holding a racket, he will enter the query ``\textit{I would like to see a girl holding racket}'' to the system. According to the query, the object ``\textit{Boy}'' in the original scene graph will be substituted with the object ``\textit{Girl}''. The target image can be retrieved given the updated scene graph. The key challenge in this process is how to modify the corresponding partial structure in the original scene graph based on understanding the natural language query.

% For example, text-based image retrieval may need to resort to an interactive retrieval process through multiple iterations of queries~\citep{Ramnath2019SceneGB,He2020SceneGM}. 

Prior effort framed this scene graph modification (SGM) task as conditional graph generation~\citep{He2020SceneGM}, where the scene graph is generated from the scratch condition on the original graph and query~\citep{You2018GraphRNNGR,Guo2019DenselyCG,Cai2020GraphTF}. However, rebuilding the entire scene graph may not be an optimal solution, as the model has to generate the partial structure of the original graph that should be unmodified. Moreover, nodes and edges of the scene graph are constructed separately in their proposed framework, which generates all the nodes first then attaches edges between generated nodes in the second pass. Such an approach may lead to the lack of the modeling capability of interactions between node prediction and edge prediction.

Instead of rebuilding the whole scene graph, we introduce a novel formulation for SGM -- incremental structure expanding (ISE), which is able to build the target graph by gradually expanding the original structure.  At each step, ISE generates the connecting edges between the existing nodes and the newly generated node, upon which the type of the new node is jointly decided. Based on the formalism, our proposed model is able to iterate between finding the relevant part in the query and reading the partially constructed scene graph, inferring more accurate and harmonious expansion decisions progressively. 
Experiments on three SGM benchmarks demonstrate the effectiveness of the proposed approach, which is able to outperform previous state-of-the-art models by large margins. To test the ability of a model under a complex scenario, we further construct a more challenging dataset from the remote sensing domain~\citep{lu2017exploring}, which has much more modification operations based on the more complicated queries compared with the existing scene graph modification datasets. 
Our key contributions are summarized as follows: 

\begin{itemize}
    \item We propose a novel formulation for scene graph modification, allowing incremental expansion of the source scene graph rather than the regeneration of the target graph.
    \item We further construct a challenging dataset that contains more complicated queries and larger scene graphs. Extensive experiments on four SGM datasets show the effectiveness of our proposed approach.
    \item  Experiments on four benchmarks demonstrate the effectiveness of our approach, which surpasses the previous state-of-the-art model by large margins. 
\end{itemize}

\section{Incremental Structure Expanding}
\label{sec:ise}
\begin{figure*}
    \centering
    \includegraphics[scale=0.67]{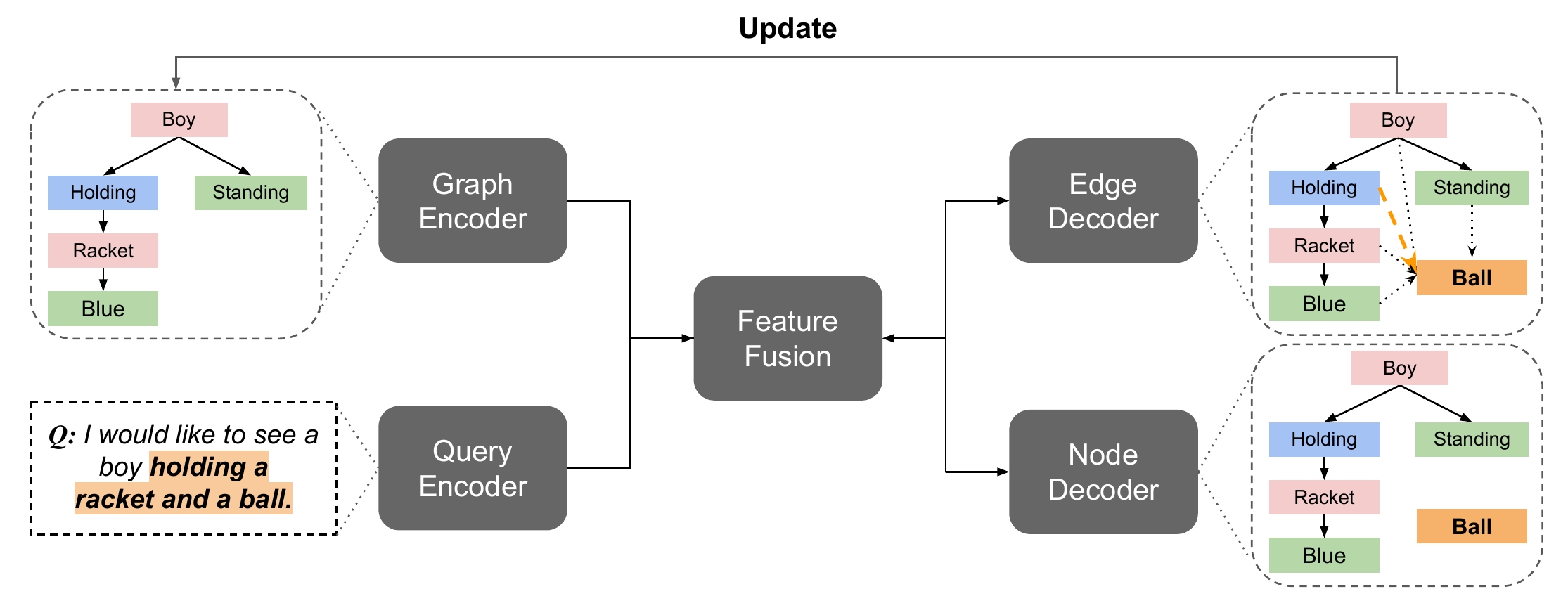}
    \caption{Overview of the model architecture. }
    \label{fig:model}
    % \vspace{-1em}
\end{figure*}

In scene graph modification, a node or multiple nodes can be inserted to, deleted from or replaced with other nodes in the scene graph. \citet{He2020SceneGM} defined the scene graph modification task as a conditional graph generation problem. Formally, given the source scene graph $G_{S}$ and the natural language query $Q$, the target scene graph $G_{T}$ is generated by maximizing the conditional probability $p(G_{T}|G_{S}, Q)$. 

Instead of generating the entire target graph $G_{T}$, we frame the task as an incremental structure expanding, which extends the source scene graph $G_{S}$ one node at a time, as well as the edges associated with the node. Such a formulation does not require the model to rebuild the unmodified structure of the source scene graph.

Under this formulation, we first define two basic operations: INSERT and DELETE. Scene graph modification can be viewed as combining and applying these two operations multiple times. Formally, given the query $Q$, a sequence of $n$ operations ${a_{1}, a_{2}, ..., a_{n}}$ are selected from a set of graph modification operations $\mathcal{A}$ = \{INSERT, DELETE\}. After applying the operations to the source scene graph $G_{s}$, the target scene graph $G_{t}$ is derived. Each operation is defined as: 

\begin{itemize}
    \item \textbf{INSERT}: A new node $o$ is added to $G_{s}$, and edges are attached between $o$ and existing nodes in $G_{s}$. As shown in Figure~\ref{fig:ise} (a), the node ``\textit{Ball}'' is added to $G_{s}$ and an edge between ``\textit{Ball}'' and ``\textit{Holding}'' is attached, according to the query ``\textit{holding a racket and a ball}''.
    \item \textbf{DELETE}: As shown in Figure~\ref{fig:ise} (b). A node $o$ is removed from $G_{s}$, as well as its associated edges. As shown in Figure~\ref{fig:ise} (b), the node ``\textit{Blue}'' is removed from $G_{s}$ and the edge between ``\textit{Racket}'' and ``\textit{Blue}'' is removed either, according to the query ``\textit{a racket not a blue}''.
\end{itemize}

Inspired by incremental parsing~\citep{Nivre2004IncrementalityID,DyerBLMS15,Cai2020AMRPV,zhang2021crowdsourcing,zhang2022identifying}, we design a data structure called extended graph $G_{I}$, which can be used to model INSERT and DELETE under the graph expansion setting. As shown in Figure~\ref{fig:ise} (a), the extended graph $G_{T}$ is identical with the target graph $G_{T}$ after applying INSERT. As for DELETE, we introduce a dummy node ``Delete'', which is attached to the node in the source graph $G_{S}$ that should be removed. For example, the dummy node ``Delete'' is attached to the node ``Blue'' in $G_{T}$. In the postprocessing stage, nodes attached with the dummy node ``Delete'' will be removed. Using this formulation, we are able to model scene graph modification by incrementally expanding the source graph $G_{S}$ to the extended graph $G_{I}$, which can be converted to the target graph $G_{T}$ without any losses. 

If the modification requires multiple operations, there will exist multiple node orderings. Take node substitution as an example, replacing a node $o_{i}$ with $o_{j}$ in $G_{s}$ can be viewed as DELETE the node $o_{i}$ first, then INSERT the node $o_{j}$, or vice versa. In practice, we impose that the DELETE operation always comes before INSERT, then the breadth-first search is used to define a deterministic node ordering.

\section{Model Architecture}
\label{sec:model}

In this section, we will present the model based on the incremental structure expanding formulation. Figure~\ref{fig:model} gives an overview of the proposed model, which consists of five components including query encoder, graph encoder, feature fusion, edge decoder and node decoder.

\paragraph{Query Encoder} This module is used to encode the query $Q$ by generating the representation of each token of it.

\paragraph{Graph Encoder} This module is used to encode the graph by generating the representation of each node of it. Note that the representations of the graph are constructed incrementally during the expanding progresses based on the updated graph of the last time step. The graph is the source graph $G_{S}$ at the first timestep.

\paragraph{Feature Fusion} this module aims to combine the representations from query and graph encoder, then served as a writable memory, which is updated based on the information from edge and node decoder during the incremental expansion. 

\paragraph{Edge Decoder} this module is used to predict the edges between the newly generated node and existing nodes of the graph, then update the memory of the feature fusion module with edge information.

\paragraph{Node Decoder} this module is used to generate a new node of the graph, then update the memory of the feature fusion module with node information.

\subsection{Query Encoder \& Graph Encoder}
For fair comparisons with the previous work~\citep{He2020SceneGM}, our query encoder and graph encoder are based on the vanilla transformer~\citep{VaswaniSPUJGKP17}, which consists of multi-head self attention (MSA) and position-wise feed-forward network (FFN) blocks. The FFN contains two layers with a ReLU non-linearity. Layer normalization (\citealt{BaKH16}) is applied before every block, and residual connections~\citep{HeZRS16} after every block. 

Formally, given an input query $Q$ with $n$ tokens, each token embedding is randomly initialized and positional encoding is added to the token embedding to retain positional information. The resulted embeddings are denotes as $\mathbf{x} = \{x_0, x_1,..., x_n\}$. Similar to BERT~\citep{DevlinCLT19}, a special token is appended to the query as $x_{0}$ for sentence encoding. Transformations in the query encoder can be denoted as:

\begin{align}
% \vspace{-2mm}
    \mathbf{x}^{l^{\prime}} = LN(MSA(\mathbf{x}^{l-1}) + \mathbf{x}^{l-1}), \\
    \mathbf{x}^{l} = LN(FFN(\mathbf{x}^{l^{\prime}}) + \mathbf{x}^{l^{\prime}}).
% \vspace{-2mm}
\end{align}

After stacking $L$ blocks, we obtained the contextualized token representations from the query encoder, denoted as $\{x_0^L, x_1^L,..., x_n^L\}$. The first vector $x_{0}$ is treated as the sentence-level representation of the query and will be used as the initial state during expansion. For clarity, we denote the vectors as $\mathbf{x}$$\in$$\mathbb{R}^{(n+1) \times d}$, where $d$ is the dimension.

As for the graph encoder, we treat the input graph as a sequence of nodes in the chronological order of when they are inserted into the graph as discussed in Section~\ref{sec:ise}. Formally, given the graph $G_{t}$ at the time step $t$, we take its node sequence $\{o_1, o_2, ..., o_{t-1}\}$ as the input. A transformer architecture is also applied to obtain the contextualized node embeddings. Notice that the contextualized representation of the graph is constructed incrementally as the expanding progress. Therefore, we apply the vanilla transformer with masked self-attention as the graph encoder, which only allows each position in the node sequence to attend to all positions up to and including that position. For brevity, we denoted the resulted contextualized node representations as $\mathbf{y}$$\in$$\mathbb{R}^{m \times d}$.

\subsection{Feature Fusion}
Unlike the conventional sequence-to-sequence model that only has one encoder, our model contains two encoders. Previous work~\citep{He2020SceneGM} proposed to use gating mechanism and cross attention to combine the representations of resulted representations from query and graph encoders. We choose to use vanilla multi-head attention mechanism~\citep{VaswaniSPUJGKP17} to fuse the features from these encoders. Formally, at each time step $t$, the feature fusion component combines the query and graph representations for gradually locating and collecting the most relevant information for the next expansion:

\begin{align}
% \vspace{-2mm}
    z_t^{l} = LN(MSA(h_t^{l-1}, \mathbf{x}) + h_t^{l-1}), \\
    z_t^{l^{\prime}} = LN(MSA(z_t^{l}, \mathbf{y}) + z_t^{l}), \\
    h_t^{l} = LN(FFN(z_t^{l^{\prime}}) + z_t^{l^{\prime}}).
% \vspace{-2mm}
\end{align}

The initial expansion state of $h_t^{0}$ is initialized with $x_{0}$. For clarity, we denote the last hidden state $h_{t}^{L}$ as $h_{t}$, which is the expansion state at the time step $t$. We now proceed to present the details of each decision stage of one expansion step.

\subsection{Edge Decoder}
At the $t$-th time step, the edge decoder takes the expansion state $h_{t}$ from the feature fusion module and the contextualized representation $\mathbf{y}$ from the graph encoder as the inputs, and predicts which nodes in the current graph should be attached to the new node. Inspired by~\citet{CaiL19a} and~\citet{Cai2020AMRPV}, we leverage multi-head attention and take the maximum over different heads as the final edge probabilities.
Formally, given $h_t$ and $\mathbf{y}$, a set of attention weights can be obtained by using multi-head attention mechanism:
$\{ \alpha_{t}^{g_{i}}\}_{i=1}^{k}$, where $k$ is the number of attention heads and $\alpha_{t}^{g_{i}}$ is the $i$-th probability vector. The probability of the edge between the new node and the node $o_{j}$ is then computed by $\alpha_{t}^{g} = max_{i}(\alpha_{t}^{g_{i}})$. Intuitively, each head is in charge of a set of possible relations (though not explicitly specified).  The maximum pooling reflects that the edge should be built once one relation is activated. 

Finally, the edge decoder passes the edge information to the feature fusion module by updating the expansion state $h_{t+1}$ as follows:

\begin{align}
    h_{t+1} = LN(MSA(h_{t},\mathbf{y}) + h_{t} ).
\end{align}

\subsection{Node Decoder}
The node decoder needs to look at the input query and determine which tokens are the most important ones. This choice is a weighted matrix that gives an attention probability between each token in the query and generated nodes in the target graph. 
% So we use a soft alignment between tokens and the new node.
Concretely, a single-head attention $\alpha_{t}^{s}$
is computed based on the state $h_t$ and the sentence representation $s_{1:n}$, where $\alpha_{t}^{s}$ denotes the attention weight of the word $w_i$ in the current time step.
This component then updates the parser state with the alignment information via the following equation:
\begin{align}
    h_{t+1} = LN(MSA(h_{t},\mathbf{x}) + h_{t} ).
\end{align}

We then compute the probability distribution of the new node through a hybrid of two channels. The new node can either be a DELETE node or a token copied from the input query. First, $h_{t}$ is fed through a $softmax$ to obtain a probability distribution over a pre-defined vocabulary, which contains the DELETE node and other dummy nodes such as $\mathsf{EOS}$. The probability of the new node is calculated as $P^{vocab} = softmax(W^{vocab}h_{t} + b^{vocab})$.

Second, we used the attention scores $\alpha_{t}^{s}$ as the probability to copy a token from the input query as a node label similar to the copy mechanism~\citep{GuLLL16,SeeLM17}. Therefore, the final prediction probability of a node $o$ is defined as:

\begin{align}
    P(o) = p_{gen} \cdot P_{vocab}(o) + p_{copy} \cdot \sum_{i \in T(c)} \alpha_{t}^{s}[i],
\end{align}

where $[i]$ indexes the $i$-th element, and
$T(c)$ are index sets of tokens respectively that have the surface form as $o$. $P(gen)$ and $P(copy)$ are the probabilities of generating  and copying a node, respectively. They are computed by using a single layer neural network with $softmax$ activation as:
\begin{align}
    [p_{gen}, p_{copy}] = softmax(W^{gate}h_{t}).
\end{align}

The whole expanding procedure is terminated if the newly generated node is the special node $\mathsf{EOS}$.

\section{Dataset Construction}
\label{sec:dataset}

\begin{table}
\centering
\scalebox{0.62}{
\begin{tabular}{l|cccc}
\toprule

Statistics & User Generated & MSCOCO & GCC & RSICD\\
%  (Train/Dev/Test) &  &  & &   \\
\midrule
Splits & 30/1/1  & 196/2/2  & 400/7/7 & 8/1/1   \\
\midrule
Avg. Source Nodes  & 2.0  & 2.9 & 3.8  & 5.9 \\
Avg. Target Nodes  & 2.0    & 2.9 & 3.7  &5.9  \\
Avg. Source Edges   & 1.0    & 1.9 & 2.8  &3.7\\
Avg. Targe Edges & 1.0     & 1.9  & 2.8   & 3.6\\
\midrule
$\mathrm{OOV}$ Nodes  & 10  & 4 & 3  & 12\\
$\mathrm{OOV}$ Edges  & 8    & 4 & 4  & 8 \\

\bottomrule
\end{tabular}}
% \vspace{-1mm}
\caption{Statistics of four SGM datasets.}
\label{tab:stats}
% \vspace{-1em}
\end{table}

Existing SGM datasets are synthetically constructed based on scene graphs from MSCOCO~\citep{Lin2014MicrosoftCC} and GCC~\citep{Sharma2018ConceptualCA}, and via crowd sourcing. To construct scene graphs, \citet{He2020SceneGM} used an in-house scene graph parser to parse a random subset of MSCOCO description data and GCC captions, thus the constructed scene graph is relatively simple. In Table \ref{tab:stats}, the average numbers of nodes and edges for each graph are limited to 2.9 and 1.9 respectively. GCC is more complicated than MSCOCO with a larger graph, but the percentage of nodes and edges from the development/test set that does not appear in the training set ($\mathrm{OOV}$ Nodes, $\mathrm{OOV}$ Edges) are still low, which will cause the model easily overfit to the dataset. To verify the generalization ability and the scalability of the model to handle more complex scene graphs, we constructed our own Scene Graph Modification dataset based on the Remote Sensing Image Captioning Dataset (RSICD)~\citep{lu2017exploring} in the remote sensing field for remote sensing image captioning task.

\begin{table*}
\centering
\scalebox{0.66}{
\begin{tabular}{lccccccccc}
\toprule
\multicolumn{1}{c}{\multirow{2}{*}{Models}} & \multicolumn{3}{c}{User Generated} & \multicolumn{3}{c}{MSCOCO} & \multicolumn{3}{c}{GCC} \\ \cmidrule(lr){2-4} \cmidrule(lr){5-7} \cmidrule(lr){8-10} 
& Node F1 & Edge F1 & GAcc & Node F1 & Edge F1 & GAcc & Node F1 & Edge F1 & GAcc \\
\midrule
CopyGraph~\citep{He2020SceneGM} & 66.17 & 31.42 & --- & 78.41 & 64.62 & --- & 79.46 & 66.32 &---\\
Text2Text~\citep{He2020SceneGM}  & 78.59 & 52.68 & 52.15 & 91.47 & 72.74 & 64.42 & --- & --- & --- \\
GRNN~\citep{You2018GraphRNNGR} & 80.68 & 57.17 & 56.75 & 80.64 & 55.76 & 50.72  & --- & --- & --- \\
DCGCN~\citep{Guo2019DenselyCG} & 79.05 & 54.23  & 52.67 & 89.08 & 72.47 & 68.89  & --- & --- & --- \\
GTran~\citep{Cai2020GraphTF} & 81.47 & 59.43  & 58.23 & 91.21 & 75.68 & 71.38  & --- & --- & ---  \\
STran~\citep{He2020SceneGM} & 83.69 & 62.10  & 60.90 & 95.40 & 86.52 & 82.97  & 93.84 & 57.68 & 52.50 \\
EGraph~\citep{weber2021extend}& 97.62 & 88.26  & 87.60 & 99.52 & 98.40 & 96.15  & 98.62 & 91.64 & 75.01 \\
\midrule

ISE  &  \textbf{98.74}\small±0.12 &  \textbf{91.37}\small±0.14  &  \textbf{89.41}\small±0.47  &  \textbf{99.68}\small±0.14 & \textbf{98.96}\small±0.21  &  \textbf{97.26}\small±0.37  & \textbf{99.53}\small±0.13  & \textbf{93.06}\small±0.22 &  \textbf{76.34}\small±0.47  \\
ISE (w/o BERT)  & 94.39\small±0.11  & 79.53\small±0.18  &  {75.72}\small±0.47  & {98.17}\small±0.13  & {97.25}\small±0.14  & {89.61}\small±0.45  &  96.91\small±0.16 & 85.50\small±0.21 &  {58.40}\small±0.56  \\
\bottomrule
\end{tabular}}
% \vspace{-1mm}
\caption{Results of User Generated, MSCOCO and GCC datasets. GAcc denotes the graph-level accuracy. Both of our models are statistically significantly outperform ($p$\textless$0.0001$) previous best-reported model~\citep{weber2021extend}.}
\label{tab:main}
% \vspace{-4mm}
\end{table*}

Inspired by the modification methods proposed by~\citet{He2020SceneGM}. First, we adopt the parser~\citep{Schuster2015GeneratingSP} to parse the caption for each graph and generate the original scene graph $\textbf{x}$. Then we define three types of graph modification operations $\mathcal{A}$ = \{INSERT, DELETE, SUBSTITUTE\}, and randomly apply them to the original scene graph to generate query ($\textbf{q}$) and modified scene graph ($\textbf{y}$). The data in RSICD consists of the triples ($\textbf{x, y, q}$).\footnote{We give three detailed operations and examples in the Appendix \ref{operations}.}

Compared with the existing SGM dataset, each graph of RSICD has more nodes and edges, with an average of 5.9 and 3.7 on the training/development/test set, which is almost twice that of User Generated and MSCOCO. In addition, the dataset comes from the field of remote sensing. Due to the large number of geographical terms, the $\mathrm{OOV}$ Nodes of the development/test sets compared with the training set reach 12\%/11\%, and the $\mathrm{OOV}$ Edges reach 8\%/8\%, which are much higher than the MSCOCO and GCC datasets. Considering the complexity of RSICD, we construct it apart from User Generated, MSCOCO and GCC to further analysis the generalization and scalability of ISE.

\section{Experiments and Analyses}
\label{sec:experiments}

\subsection{Data}
We evaluated our model on four benchmarks, including User Generated, MSCOCO and GCC proposed by~\citet{He2020SceneGM}, and RSICD dataset proposed in this work. MSCOCO, GCC and RSICD are constructed synthetically from publicly available datasets~\citep{Lin2014MicrosoftCC, SoricutDSG18, lu2017exploring}, while the User Generated dataset is created via crowd sourcing. Detailed statistics of datasets are shown in Table~\ref{tab:stats}.

\subsection{Setup}

For fair comparisons, we used the same data splits for User Generated, MSCOCO and GCC datasets as in ~\citet{weber2021extend}. For RSICD, we randomly split the data into 8K/1K/1K for training/development/test. Following~\citet{weber2021extend}, we use three automatic metrics for the evaluation, including node-level and edge-level F1 score, and graph-level accuracy. Graph-level accuracy is computed based on exact string match, which requires the generated scene graph to be identical to the target scene graph for a correct prediction. We reported the mean score and standard deviation by using 5 models from independent runs. We refer to the Appendix \ref{Hyper-parameters} for the detailed implementation.

\subsection{Baselines}
For comprehensive comparisons, we include six baselines as follows. Except for the CopyGraph, all of them aim to rebuild the target scene graph. 

\paragraph{CopyGraph} This baseline directly copies the source scene graph as the target scene graph, which can be viewed as the lower bound.

\paragraph{Text2Text} This baseline is introduced by~\citet{He2020SceneGM}. They used the standard sequence-to-sequence architecture by linearizing the scene graph based on depth-first search. 

\paragraph{GRNN} Graph RNN~\citep{You2018GraphRNNGR} is used as the graph encoder and edge decoder. Specifically, the edges are represented by an adjacency matrix, which is then generated in an auto-regressive manner. Both the query encoder and node decoder are based on Gated Recurrent Units~\citep{ChoMGBBSB14}.

\paragraph{DCGCN} Densely-Connected Graph Convolutional Networks ~\citep{Guo2019DenselyCG} are used as the graph encoder. Other components are kept the same as the GRNN.

\paragraph{GTran} Graph Transformer~\citep{Cai2020GraphTF} is used as the graph encoder, while other modules are the same as GRNN and DCGCN.

\paragraph{STran} The sparsely-connected transformer~\citep{He2020SceneGM} is used to encode the source graph. In addition, a cross-attention mechanism is applied to fuse the features from graph encoder and query encoder. Node decoder and edge decoder are the same as GRNN.

\paragraph{EGraph} This is the state-of-the-art model on graph modification task. Concretely, \citet{weber2021extend} considerably increases performance on the graph modification by phrasing it as a sequence labelling task.

\subsection{Main Results}
According to Table~\ref{tab:main}, our proposed approach (ISE) significantly outperforms the state-of-the-art model~\citep{weber2021extend} on three datasets. Specifically, ISE outperforms EGraph 1.81, 1.11 and 1.33 percentage points in terms of graph accuracy on User Generated, MSCOCO and GCC datasets, respectively. We observe that the improvement is especially prominent on the User Generated dataset, which is 
more challenging than the other two synthetic datasets in terms of the diversity in graph semantics and natural language expressions. All baseline models suffer from performance degradation as it is much harder to rebuild the entire target scene graph on this dataset. On the other hand, ISE constructs the target scene graph by incrementally expanding the source scene graph without changing the unmodified structure. We believe this formulation is able to effectively cope with this difficulty.

We also observe that both EGraph and ISE achieve lower graph accuracy on the GCC dataset. The main reason is the difficulty of predicting the correct edges between generated nodes. For example, EGraph achieves 98.62 Node F1 score on GCC, higher than 97.62 Node F1 score on the User Generated dataset. However, EGraph only achieves 75.01 Edge F1 score on GCC, while it can attain 88.26 Edge F1 score on User Generated. Our proposed model has larger improvements upon EGraph in terms of Edge F1 score on the same dataset (93.06 vs. 91.64). We attribute this stronger improvement to iterations between nodes prediction and edge prediction, which allows more accurate and harmonious expansion decisions progressively. On the other hand, EGraph predicts nodes and edges at two independent stages. Such an approach may lead to the lack of the modeling capability of interactions between node prediction and edge prediction.

\begin{table}
\centering
\scalebox{0.68}{
\begin{tabular}{lccccccccc}
\toprule
\multicolumn{1}{c}{\multirow{2}{*}{Models}} & \multicolumn{3}{c}{RSICD} \\ \cmidrule(lr){2-4}  
& Node F1 & Edge F1 & GAcc \\
\midrule
CopyGraph & 66.35 & 58.68 & --- \\
EGraph~\citep{weber2021extend} & 72.09\small±0.12 & 53.96\small±0.31  & 23.93\small±0.74 \\
\midrule
% ISE  & 77.37\small±0.16  & 65.53\small±0.28  &  {40.73}\small±0.67    \\
ISE  &  \textbf{81.78}\small±0.13 &  \textbf{67.01}\small±0.25 &  \textbf{44.20}\small±0.59  \\
\bottomrule
\end{tabular}}
% \vspace{-2mm}
\caption{Results of the RSICD dataset. Results of STran is
reproduced from the implementation of~\citet{weber2021extend}. Both of our models are statistically significantly outperform ($p$\textless$0.0001$) previous best-reported model~\citep{weber2021extend}.}
\label{tab:rs}
% \vspace{-4mm}
\end{table}

We further compare our model with EGraph on the newly constructed dataset RSICD as shown in Table~\ref{tab:rs}. ISE is able to achieve a graph accuracy of 44.20\% and improves upon the EGraph model by 21 percentage points. However, the graph accuracy of all the models is much lower than the one attained on the previous three SGM datasets. One reason is that RSICD has more complex queries paired with larger scene graph, which brings a challenge to existing models. The RSICD dataset also suffers from the data sparsity issue where many words (39\%) and nodes (42\%) only appear once in the training data. Incorrect node prediction will further propagate the errors to edge prediction. Our iterative node and edge prediction paradigm help to alleviate this issue. Specifically, ISE only outperforms EGraph 9.69 percentage points on Node F1 score, while the improvement on Edge F1 score is 13.05\%. Therefore, ISE is able to achieve a higher accuracy. In order to further address this data sparsity issue, one potential solution is transfer learning, where the model is pretrained on User Generated dataset first then fine-tuned on RSICD. However, this approach may suffer from a domain-shift problem, as RSICD is constructed based on the remote sensing domain. We leave this direction as future works.

\subsection{Analysis and Discussion}
In this section, we provided a fine-grained analysis of our proposed model. We reported all the results on the development set by using the ISE model without contextualized embeddings from BERT.

\begin{table}
\centering
\scalebox{0.75}{
\begin{tabular}{ll|ccc}
\toprule
\multicolumn{2}{c}{Datast/Model} & Node F1 & Edge F1 & GAcc\\
\midrule
\multirow{3}{*}{User Generated} & ISE & 94.58 & 79.61 & 76.23   \\
  & ISE Rebuild & 82.97 & 66.74 & 63.47 \\
  & ISE - Copy & 88.38 & 74.95  & 71.64 \\
  \midrule
\multirow{3}{*}{MSCOCO} & ISE & 98.60 & 97.99 & 92.24   \\
  & ISE Rebuild & 92.90 & 87.61  & 83.23 \\
  & ISE - Copy & 95.58 & 91.87 & 88.09 \\
  \midrule
\multirow{3}{*}{GCC} & ISE   & 96.87 & 85.50 & 58.90  \\
  & ISE Rebuild & 89.48 & 62.59  & 51.67 \\
  & ISE - Copy & 92.74 & 76.89 & 53.91 \\
\bottomrule
\end{tabular}}
% \vspace{-2mm}
\caption{ An ablation study for ISE. Rebuild denotes that we regenerate the scene graph rather than extend it. - Copy denotes model without using the copy mechanism}
\label{tab:ablation}
% \vspace{-2mm}
\end{table}

\paragraph{Ablation Study} 
As shown in Table~\ref{tab:ablation}, we examine the contributions of two main components used in our model. The first one is the incremental structure expanding. We use the same model architecture but try to rebuild the target scene graph similar to previous efforts. We can observe significant drops on three SGM datasets, which further confirms the effectiveness of the extending strategy. The second one is the copy mechanism, which directly copies the token from the query as nodes in the target scene graph. It plays a significant role in predicting nodes especially when the training data is limited (User Generated).

\begin{table}
\centering
\scalebox{0.85}{
\begin{tabular}{ll|ccc}
\toprule
\multicolumn{2}{c}{{\% of Training Set}} & Node F1 & Edge F1 & GAcc\\
\midrule
\multirow{2}{*}{20\%} & STran   & 87.92 & 71.35 & 68.15  \\
  & ISE & \textbf{95.46} & \textbf{92.22} & \textbf{79.12} \\
  \midrule
\multirow{2}{*}{40\%} & STran   & 93.94 & 81.11 & 78.55  \\
  & ISE & \textbf{97.50} & \textbf{96.12} & \textbf{88.64} \\
  \midrule
\multirow{2}{*}{60\%} & STran   & 95.32 & 82.70 & 80.65  \\
  & ISE & \textbf{98.09} & \textbf{97.17} & \textbf{89.29} \\
  \midrule
\multirow{2}{*}{80\%} & STran   & 95.92 & 86.36 & 83.90  \\
  & ISE & \textbf{98.37} & \textbf{97.48} & \textbf{90.69} \\
  \midrule
\multirow{2}{*}{100\%} & STran   & 96.24 & 87.88 & 85.20  \\
  & ISE & \textbf{98.60} & \textbf{97.99} & \textbf{92.24} \\
\bottomrule
\end{tabular}}
% \vspace{-2mm}
\caption{Comparison of STran and ISE against different training data sizes on the dev set of MSCOCO. Results of STran are
reproduced from ~\citet{He2020SceneGM}.}
\label{tab:size}
% \vspace{-4mm}
\end{table}

\paragraph{Performance against Training Data Size} Table~\ref{tab:size} shows the performance of STran and ISE against different training settings on MSCOCO dataset. We considered four training settings (20\%, 40\%, 60\%, 80\%, 100\% training data). ISE consistently outperforms STran under the same amount of training data. When the size of training data decreases, we can observe that the performance gap becomes more obvious. Particularly, using 40\% of the training data, ISE is able to achieve a graph accuracy of 88.64\%, higher than STran trained on the whole dataset. These results demonstrate that our model is more effective in terms of using training resources and more robust when the training data is limited.

\begin{table}
\centering
\scalebox{0.9}{
\begin{tabular}{ll|ccc}
\toprule
\multicolumn{2}{c}{{Query Length}} & Node F1 & Edge F1 &Graph Acc\\
\midrule
\multirow{2}{*}{\textless 5 } & STran   & 51.68 & 91.38 & 40.98  \\
  & ISE & \textbf{96.02} & \textbf{92.36} & \textbf{57.38} \\
  \midrule
\multirow{2}{*}{5$\sim$10} & STran   & 92.73 & 57.01 & 50.39  \\
  & ISE & \textbf{97.32} & \textbf{86.84} & \textbf{60.18} \\
  \midrule
\multirow{2}{*}{${\geq}$10} & STran   & 91.38 & 51.68 & 40.98  \\
  & ISE & \textbf{98.42} & \textbf{84.62} & \textbf{58.64} \\
\bottomrule
\end{tabular}}
% \vspace{-1mm}
\caption{Comparison of STran and ISE against different lengths of queries.}
\label{tab:query}
% \vspace{-4mm}
\end{table}

\paragraph{Performance against Query Length} Table~\ref{tab:query} shows the results of STran and ISE under different query lengths on GCC dataset. We partitioned the sentence length into three classes (\textless5, [5, 10), $\geq$10). In general, ISE outperforms STran against various sentence lengths. When the length of the query increases, we can observe that the performance gap becomes more obvious in terms of graph accuracy. Intuitively, with the increase of the query length, it is more challenging for the model to comprehend the sentence. This suggests that ISE is able to handle more complex instructions.

\paragraph{Performance against Graph Size}
Table~\ref{tab:graph} shows the results of STran and ISE against different target scene graph sizes on GCC dataset. We partitioned the scene into three classes (\textless5, [5, 10), $\geq$10).  Based on the formulation of extending the source scene graph, our model is required to deal with larger graphs. For example, deleting a node in the scene graph becomes adding a special ``Delete'' node in the extended graph. However, ISE is able to consistently outperform STran against various target graph sizes, even when the target scene graph is large. This result suggests the superiority of the proposed formulation.\footnote{We give an error analysis in the Appendix \ref{error}.} 
\begin{table}
\centering
\scalebox{0.85}{
\begin{tabular}{ll|ccc}
\toprule
\multicolumn{2}{c}{{Graph Size}} & Node F1 & Edge F1 &Graph Acc\\
\midrule
\multirow{2}{*}{\textless 5 } & STran   & 94.04 & 62.77 & 58.74  \\
  & ISE & \textbf{96.73} & \textbf{81.25} & \textbf{62.54} \\
\midrule
\multirow{2}{*}{5$\sim$10 } & STran   & 91.61 & 51.40 & 37.73  \\
  & ISE & \textbf{97.22} & \textbf{88.82} & \textbf{49.06} \\
\midrule
\multirow{2}{*}{${\geq}$10 } & STran   & 79.12 & 30.13 & 24.62  \\
  & ISE & \textbf{95.44} & \textbf{90.95} & \textbf{35.38} \\
\bottomrule
\end{tabular}}
% \vspace{-1mm}
\caption{Comparison of STran and ISE against different target scene graph sizes.}
\label{tab:graph}
% \vspace{-4mm}
\end{table}

\begin{figure*}
    \centering
    \includegraphics[scale=0.26]{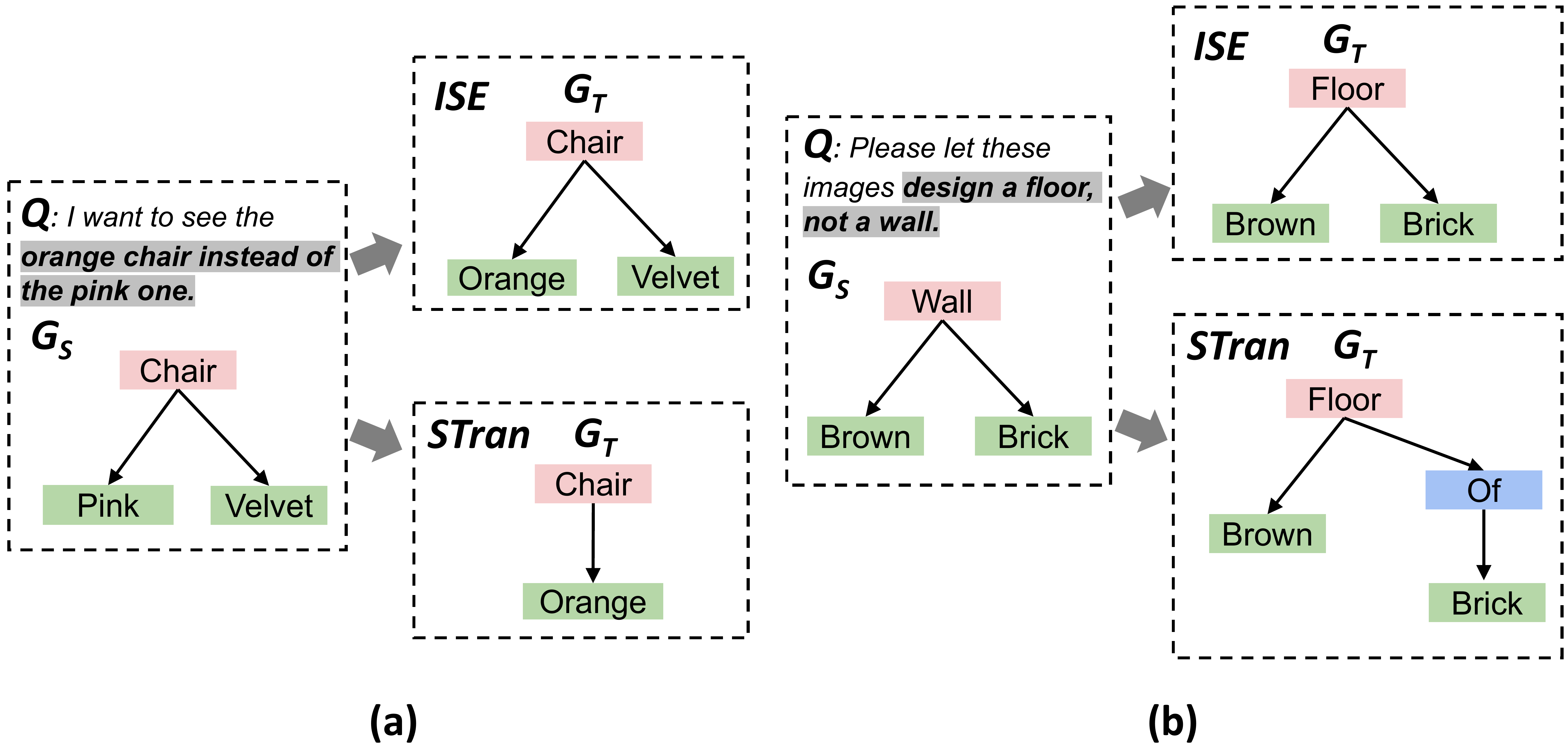}
    % \vspace{-1em}
    \caption{Two cases of STran and ISE for scene graph modification on User Generated. $Q$ denotes the textual query, $G_{S}$ denotes the source scene graph, $G_{T}$ denotes the target scene graph generated by STran and ISE.}
    \label{fig:case}
    % \vspace{-1em}
\end{figure*}

\paragraph{Case Study}
\label{case}

We give two cases in Figure \ref{fig:case}. STran generates scene graph from the scratch conditioned on the original graph and query may lead to the lack of the modeling capability of interactions between node prediction and edge prediction. For example, in Figure \ref{fig:case} (a), STran omitted the attribute: ``Velvet'' during the node prediction. In addition, during the edge prediction, STran redundantly generated the relation: ``Of'' in Figure \ref{fig:case} (b). However, these structures do not need to be modified in the source scene graph. ISE can infer more accurate target graph by incrementally expanding the source graph without changing the unmodified structure. 
\section{Related Work}
\label{sec:related}

We refer to the Appendix \ref{scene graph} for the detailed related work of scene graph. Scene graph builds a bridge between image domain and text domain. Vision and natural language are all tremendously promoted by studying into scene graphs. Recently, scene graph modification becomes an emerging research
direction. \citet{Chen2020GraphED} proposed a framework based on scene graph editing for text-based image retrieval.
On the other hand, \citet{He2020SceneGM} took the scene graph and the textual query as inputs and modified the source graph according to the query. They viewed the task as conditional graph generation, which is further decomposed into node prediction and edge prediction. For node prediction, all the nodes in the target scene graph is generated based on a graph-to-sequence model with dual encoder~\citep{Song2018AGM,Beck2018GraphtoSequenceLU,ZhangGTLCLB20}, then a graph RNN is adopted to predict the edges between generated nodes~\citep{You2018GraphRNNGR}. More recently, \citet{weber2021extend} developed an alternative formulation of this problem in which they model the modification as an auto-regressive sequence labelling task.
% to form the entire scene graph.
% \input{tables/NodeLength}

Instead of rebuilding the entire target graph, we framed the scene graph modification task as incremental graph expansion. This formulation is related to incremental parsing, where a sentence is scanned from left-to-right and the structured is built incrementally by inserting a node or attaching an edge. Incremental parsers are widely used in semantic parsing~\citep{ZhouXUQLG16,ChengRSL17,GuoL18, Naseem2019RewardingST,liu2022semantic} and syntactic parsing~\citep{HuangS10,DyerBLMS15,LiuZ17a}, as they are computationally efficient, and can use machine learning to predict actions based on partially generated structures. Our feature fusion module can be viewed as the parser state as it carries the structural information and serves as a writable memory during the expansion step. Unlike \citet{weber2021extend} linearize the scene graph and label it in an auto-regressive manner, our model iterates between finding the relevant part in the query and reading the partially constructed scene graph, inferring more accurate and harmonious expansion decisions progressively.

\section{Conclusion}

In this paper, we designed a novel formulation for scene graph modification, which allows us to incrementally expand the source scene graph instead of rebuilding the entire graph. Based on the formalism, we further propose a model that is able to leverage the mutual causalities between node prediction and edge prediction. Experiments on three SGM benchmarks demonstrate the effectiveness.
% of our proposed approach. 
To test our model under a complex scenario, we constructed a more challenging dataset from the remote sensing domain, which has more modification operations based on the more complicated queries compared with existing SGM datasets. For future work, we would like to explore how to integrate the model into the text-based image retrieval task.
% ~\citep{Chen2020GraphED}.

% \section{Acknowledgement}
% We thank the reviewers for their valuable comments. The work was supported by the National Key Research and Development Program of China (No. 2019YFB1704003), the National Nature Science Foundation of China (No. 62021002 and No. 71690231), NSF under grants III-1763325, III-1909323, III-2106758, SaTC-1930941, Tsinghua BNRist and Beijing Key Laboratory of Industrial Bigdata System and Application.
% \input{subfiles/7_ethics}
\section{Acknowledgement}
We thank the reviewers for their valuable comments. The work was supported by the National Key Research and Development Program of China (No. 2019YFB1704003), the National Nature Science Foundation of China (No. 62021002 and No. 71690231), NSF under grants III-1763325, III-1909323, III-2106758, SaTC-1930941, Tsinghua BNRist and Beijing Key Laboratory of Industrial Bigdata System and Application.

\bibliography{custom}
\bibliographystyle{acl_natbib}

\newpage
\appendix
\clearpage
\section{Appendix}

\subsection{Operations in RSICD}

\begin{table}[t!]
\centering
\resizebox{1.02\linewidth}{!}{
\begin{tabular}{l}
\toprule
\begin{tabular}[c]{@{}l@{}}\textbf{Graph Modification Operation:} \textbf{\texttt{DELETE}}
\\ 
\textbf{Original Scene Graph:} Some {\color{red}\textit{trees}} are in a medium residential area.
\\
\textbf{Query:} Remove {\color{red}\textit{trees}}.
\\\textbf{Modified Scene Graph:} Some are in a medium residential area.
\end{tabular}     \\
\\

\begin{tabular}[c]{@{}l@{}}\textbf{Graph Modification Operation:} \textbf{\texttt{INSERT}}
\\ 
\textbf{Original Scene Graph:} A bridge built on a river.
\\
\textbf{Query:} Show me a {\color{red}\textit{red}} bridge.
\\\textbf{Modified Scene Graph:}A {\color{red}\textit{red}} bridge built on a river.

\end{tabular}     \\
\\

\begin{tabular}[c]{@{}l@{}}\textbf{Graph Modification Operation:} \textbf{\texttt{SUBSTITUTE}}
\\ 
\textbf{Original Scene Graph:} Some gray and {\color{green}\textit{green}} mountains are together.
\\
\textbf{Query:} I prefer {\color{red}\textit{red}} to {\color{green}\textit{green}}, modify {\color{red}\textit{red}} to {\color{green}\textit{green}}.
\\\textbf{Modified Scene Graph:} Some gray and {\color{red}\textit{red}} mountains are together.
\end{tabular}     \\
\bottomrule
\end{tabular}}
\caption{Examples on the three types of graph modification operations $\mathcal{A}$ = \{INSERT, DELETE, SUBSTITUTE\} }\label{tab:RSICD_case_study}
% \vspace{-6mm}
\end{table}
\label{operations}
We introduce three operations in RSICD in details:
\begin{itemize}
    \item \textbf{\texttt{DELETE}}: The original scece graph is $\textbf{x}$. We randomly select a node $\textbf{o}$ in $\textbf{x}$, and delete it both with related edges. The deleted graph is defined as $\textbf{y}$. We choose a random sentence from the \textit{DELETE Template}~\citep{manuvinakurike-etal-2018-edit}, for example, `` I do not want \textbf{**}.'' We replace \textbf{**} with $\textbf{o}$ to get modification operation $\textbf{q}$.
    \item \textbf{\texttt{INSERT}}: It is the reverse process of \textbf{\texttt{DELETE}}. The graph before deleting the node is regarded as $\textbf{y}$, and the corresponding graph after deletion is treated as $\textbf{x}$. The modification operation is randomly selected from the \textit{INSERT Template}~\citep{manuvinakurike-etal-2018-edit}, for example, `` Show me \textbf{**}.'' We replace \textbf{**} with $\textbf{o}$ to obtain query $\textbf{q}$.
    \item \textbf{\texttt{SUBSTITUTE}}: We randomly select a node $\textbf{o}$, use the AllenNLP toolkit~\citep{gardner-etal-2018-allennlp} to find the three most similar semantics nodes compared with $\textbf{o}$. We randomly choose a node $\textbf{m}$, and select a sentence from the \textit{SUBSTITUTE Template}~\citep{manuvinakurike-etal-2018-edit}, for example, `` I prefer \textbf{@@} to \textbf{**}, modify \textbf{**} to \textbf{@@}.'' We replace \textbf{**} and \textbf{@@} with $\textbf{o}$ and $\textbf{m}$, and get modification operation $\textbf{q}$. Note that SUBSTITUTE operation could be viewed as DELETE the node $\textbf{o}$ first and then INSERT the node $\textbf{m}$, or vice versa.
\end{itemize}

In Table \ref{tab:RSICD_case_study}, we give the simple examples in RSICD to better understand three types of graph modification operations.
% $\mathcal{A}$ = \{INSERT, DELETE, SUBSTITUTE\}.

\subsection{Implementation Details}
\label{Hyper-parameters}
Hyper-parameters of the model are tuned on the development set. All transformer~\citep{VaswaniSPUJGKP17} layers share the same hyper-parameter settings. Following~\citet{He2020SceneGM}, we randomly initialized the word and node embeddings. We also report results with contextualized embeddings from BERT~\citep{DevlinCLT19}. Specifically, we used the BERT-base-uncased implemented by~\citep{wolf-etal-2020-transformers}. The parameters in BERT are fixed during training. To mitigate over-fitting, we apply dropout~\citep{SrivastavaHKSS14} with the drop rate 0.2 between different layers.  Following~\citet{Cai2020AMRPV}, we use a special UNK token to replace the out-of-vocabulary lemmas of the input query and remove the UNK token in the generated graph. Parameter optimization is performed with the ADAM optimizer~\citep{KingmaB14} with $\beta_{1}$ = 0.9 and $\beta_{2}$ = 0.999. The learning rate schedule is similar to that in~\citet{VaswaniSPUJGKP17}, where warm-up steps being set to 2K. We used early stopping on the development set for choosing the best model. 
Please refer to Table \ref{tab:hyper-parameters} for the detailed hyper-parameters settings for ISE.

\begin{table}
\centering
\scalebox{0.9}{
\begin{tabular}{lr}
\toprule
\multicolumn{2}{l}{\textbf{Embeddings}}\\
\midrule
concept  & 300 \\
word & 300   \\
relation  & 100\\
\midrule
\multicolumn{2}{l}{\textbf{Query Encoder}}\\
transformer layers & 4   \\
\midrule
\multicolumn{2}{l}{\textbf{Graph Encoder}}\\
transformer layers & 2   \\
\midrule
\multicolumn{2}{l}{\textbf{Feature  Fusion}}\\
heads & 8   \\
hidden size & 512   \\
feed-forward hidden size & 1024   \\
\midrule
\multicolumn{2}{l}{\textbf{Node Decoder/ Edge Decoder}}\\
heads & 8   \\
feed-forward hidden size & 1024   \\

\bottomrule
\end{tabular}}

\caption{Hyper-parameters settings for ISE.}
\label{tab:hyper-parameters}

\end{table}

\begin{figure*}[ht]
    \centering
    \includegraphics[scale=0.26]{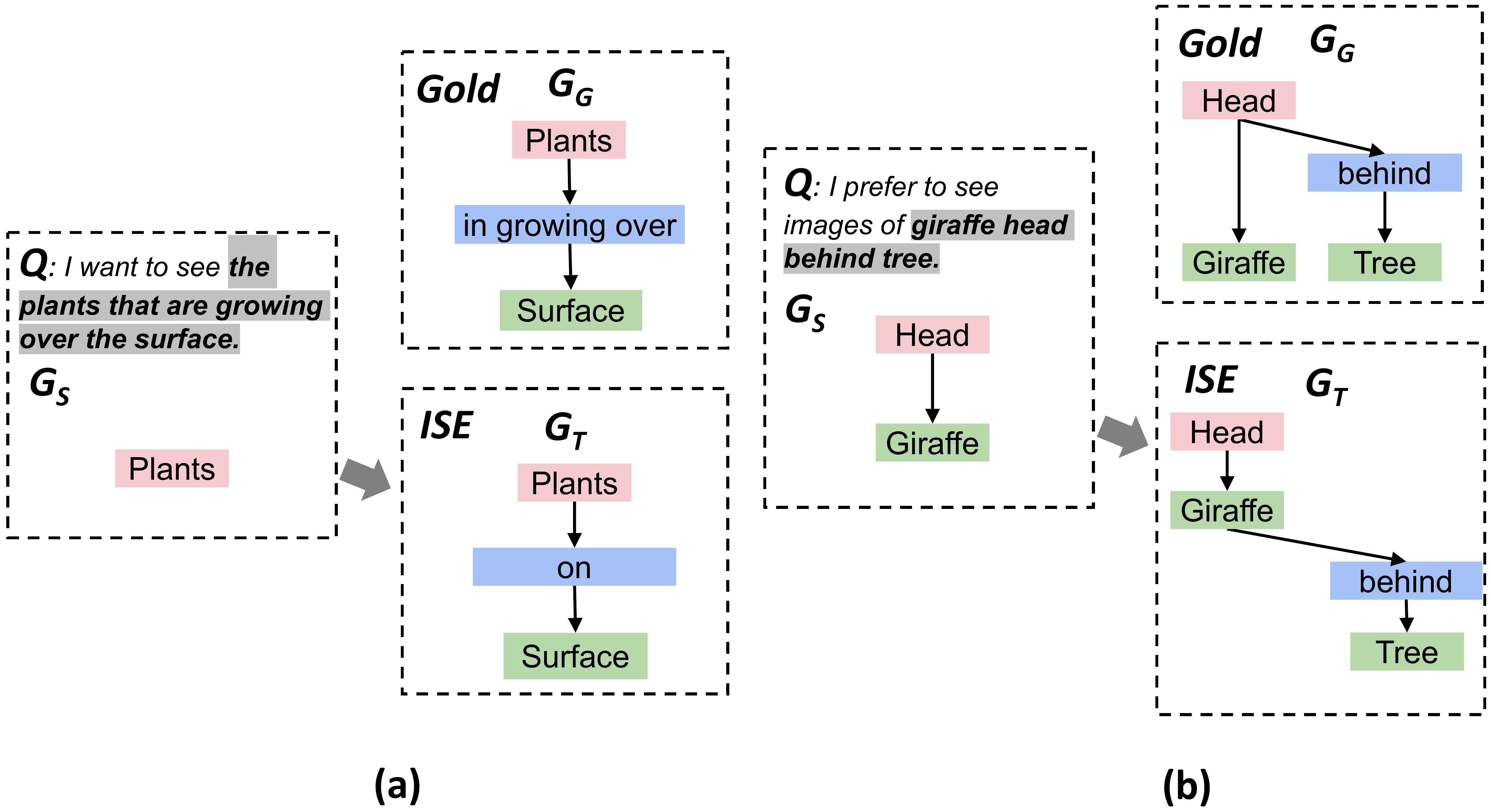}
    % \vspace{-2em}
    \caption{Two errors of ISE for scene graph modification on User Generated. $Q$ denotes the textual query, $G_{S}$ denotes the source scene graph, $G_{T}$ denotes the target scene graph generated by ISE. $G_{G}$ denotes the gold target scene graph.}
    \label{fig:error}
    % \vspace{-1em}
\end{figure*}

\subsection{Scene Graph and Application}
\label{scene graph}
Deep learning has significantly promoted the advancement of computer vision~\citep{liang2017deep, ren2021comprehensive}. Simple visual understanding tasks such as object detection and recognition are no longer sufficient. To depict the relationship between objects in the scene as a driving force, higher-level visual understanding and reasoning skills are frequently necessary. Scene graphs were created specifically to address this issue. Scene graph was first proposed by~\citet{Johnson2015ImageRU} for image retrieval, which describes objects, their attributes, and relationships in images with a graph. A complete scene graph could represent the semantics of a dataset's scenes, not just a single image or video; additionally, it contains powerful representations that encode 2D/3D images~\citep{Johnson2015ImageRU,armeni20193d}, and videos~\citep{qi2018scene,wang2020storytelling} into their abstract semantic elements. Scene graph is beneficial for various downstream tasks, such as information extraction \cite{hu2020selfore,hu2021semi,hu2021gradient,liu2022hierarchical}, natural language summarization \cite{liu2022psp}, and natural language inference \cite{li2022pair}.

Following the graph representation paradigm, different methods have been proposed to generate scene graphs from images~\citep{XuZCF17,WangLZY18, ZellersYTC18}. Many cross-modal tasks that require understanding and reasoning on image and text are able to benefit from incorporating scene graphs, such as visual question answering~\citep{TeneyLH17,shi2019explainable}, grounding referring expressions~\citep{wang19}, image captioning~\citep{YangTZC19,yao2018exploring}, and image retrieval~\citep{Wang2020CrossmodalSG,Schroeder2020StructuredQI}.

\subsection{Error Analysis} 
\label{error}
We give two wrong scene graphs generated by ISE in Figure \ref{fig:error}. We can observe in Figure \ref{fig:error} (a) that although ISE successfully predicts the need to insert a relation between object ``Plants'' and attribute ``Surface'', since the User Generated dataset contains a total of 2078 relations and the relations have serious long-tail effects. It is difficult for ISE to learn sparseness relations with few occurrences, leading to incorrectly predicting relation ``in growing over'' as ``on''. We attempt to address the long-tail effects of relations in future work.
Since a node can be attached to multiple nodes, when Edge Decoder determines which nodes in the current graph should be attached to the new node, a common error is predicting the wrong node that needs to be attached. As shown in Figure \ref{fig:error} (b), ISE incorrectly connects relation ``behind'' between ``Giraffe'' and ``Tree'' instead of ``Head'' and ``Tree''.

\end{document}